%% file: Mitigating_Artifacts_in_NLI_to_Transfer_Models_across_Datasets_arxiv.tex
\documentclass[11pt,a4paper]{article}
\usepackage[table]{xcolor}
\usepackage[hyperref]{acl2019}
\usepackage{times}
\usepackage{latexsym}

\input{math_commands.tex}

\usepackage[table]{xcolor}
\usepackage{pgfplotstable}
\usepackage{diagbox}

\pgfplotstableset{
color cells/.style={
col sep=comma,
string type,
postproc cell content/.code={
\pgfkeysalso{@cell content=\rule{0cm}{2.4ex}\cellcolor{black!##1}\pgfmathtruncatemacro\number{##1}\ifnum\number>50\color{white}\fi##1}
},
columns/x/.style={
column name={\backslashbox{$\beta$}{$\alpha$}},
postproc cell content/.code={}
}
}
}

\usepackage{soul,color}
\usepackage{hyperref}
\usepackage{url}

\usepackage{subcaption}
\usepackage{amsmath}
\usepackage{comment}
\usepackage{booktabs}
\usepackage{tabularx}
\usepackage{dcolumn}
\newcolumntype{d}[1]{D{.}{\cdot}{#1} }

\usepackage{adjustbox}

\usepackage{dcolumn}
\newcolumntype{H}{>{\setbox0=\hbox\bgroup}c<{\egroup}@{}}

\usepackage[disable]{todonotes}
\newcommand{\adam}[1]{\todo[color=blue!40]{\tiny Adam: #1}}

\newcommand{\yb}[1]{\todo[color=red!40]{\tiny Yonatan: #1}}

\usepackage{xspace}
\newcommand{\hypoth}{H\xspace}
\newcommand{\premise}{P\xspace}
\newcommand{\enc}{g}

\newcommand{\True}{\textsc{True}\xspace}
\newcommand{\False}{\textsc{False}\xspace}
\newcommand{\HOC}{Method 1\xspace}
\newcommand{\NS}{Method 2\xspace} 
\newcommand{\NSlong}{Method 2\xspace} 
\newcommand{\HOClong}{Method 1\xspace}

\renewcommand{\mid}{\,|\,}

\newcommand{\newfigref}[1]{Figure~\ref{#1}}
\newcommand{\tabref}[1]{Table~\ref{#1}}
\newcommand{\appref}[1]{Appendix~\ref{#1}}

\usepackage{caption}

\usepackage{paralist}
\usepackage{enumitem}
\usepackage{lipsum,multicol}

\usepackage{soul}

\title{
Don't Take the Premise for Granted: \\
Mitigating Artifacts in Natural Language Inference
}

\author{Yonatan Belinkov$^{13}$\thanks{$^*$ Equal contribution} \hspace{1em} Adam Poliak$^{2*}$ \\
\textbf{\hspace{1em} Stuart M. Shieber$^1$ \hspace{1em} Benjamin Van Durme$^2$ \hspace{1em} Alexander M. Rush$^1$}\\
$^1$Harvard University \hspace{0.9em} 
$^2$Johns Hopkins University \hspace{0.9em} 
$^3$Massachusetts Institute of Technology \\
\texttt{\{belinkov,shieber,srush\}@seas.harvard.edu} \\ 
\texttt{\{azpoliak,vandurme\}@cs.jhu.edu}
}

\newcommand{\cameraready}[1]{\xspace}
\renewcommand{\cameraready}[1]{\xspace}
\newcommand{\sout}[1]{}
\renewcommand{\sout}[1]{}
\newcommand{\adversarialDescription}[1]{}
\renewcommand{\adversarialDescription}[1]{}
\newcommand{\jointDescription}[1]{}
\renewcommand{\jointDescription}[1]{}

\aclfinalcopy 
\def\aclpaperid{804} 
\begin{document}

\maketitle

\begin{abstract}
Natural Language Inference (NLI) datasets often contain hypothesis-only biases---artifacts
that allow models to achieve non-trivial performance without learning whether a premise entails a hypothesis.
We propose two probabilistic methods to build models that are more robust to such biases
and better transfer across datasets.
In contrast to standard approaches to NLI, 
our methods predict the probability of a premise given
a hypothesis and NLI label, discouraging models from ignoring the premise.
We evaluate our methods on synthetic and existing NLI datasets
by training
on datasets containing biases and testing on 
datasets containing no (or different) hypothesis-only biases. 
Our results indicate that these methods can make NLI models more robust to dataset-specific artifacts,
transferring better than a baseline architecture in $9$ out of $12$ NLI datasets. 
Additionally, we provide
an extensive analysis of the interplay of our methods with known biases in NLI datasets, as well as the effects of encouraging models to ignore biases and fine-tuning 
on target datasets.
\footnote{Our code is available at \url{https://github.com/azpoliak/robust-nli}.}

\end{abstract}

\section{Introduction}

Natural Language Inference (NLI) 
is often used to gauge
a model's ability to understand 
a relationship
between two texts~\citep{fracas,dagan2006pascal}.
In NLI, a model is tasked with determining whether a hypothesis 
(\textit{a woman is sleeping}) would likely be inferred from a premise (\textit{a woman is talking on the phone}).\footnote{This hypothesis contradicts the premise and would likely not be inferred.}
The development
of new large-scale datasets has led to a flurry of
various neural network architectures
for solving
NLI. 
However,
recent work has found that
many NLI datasets contain biases, or annotation artifacts, i.e., features present in hypotheses that enable
models to perform surprisingly well using only the hypothesis, 
without learning the relationship between two texts~\citep{gururangan-EtAl:2018:N18-2,poliak-EtAl:2018:S18-2,1804.08117}.\footnote{We
use 
\textit{artifacts} and \textit{biases} interchangeably.}
For instance, in some datasets, negation words like ``not'' and ``nobody'' are often associated with a relationship of contradiction.
As a ramification of such biases,
models may not generalize well to
other datasets that contain different or no such biases.

\begin{figure*}[t]
\centering
\begin{subfigure}[b]{0.32\linewidth}
\centering
\includegraphics[height=3.5cm]{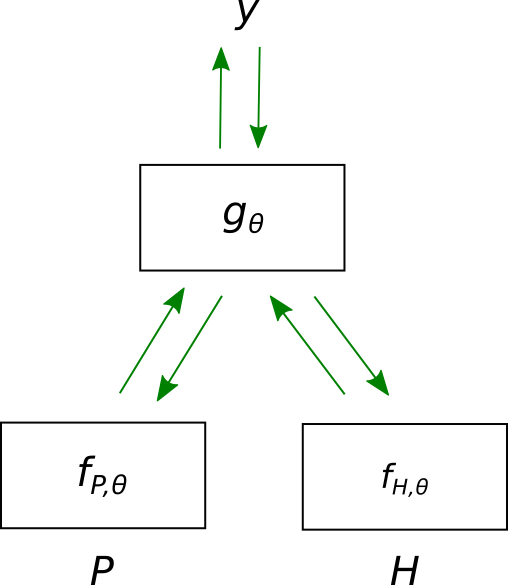}
\caption{Baseline}
\label{fig:arch-baseline}
\end{subfigure} \hfill
\begin{subfigure}[b]{0.32\linewidth}
\centering
\includegraphics[height=3.5cm]{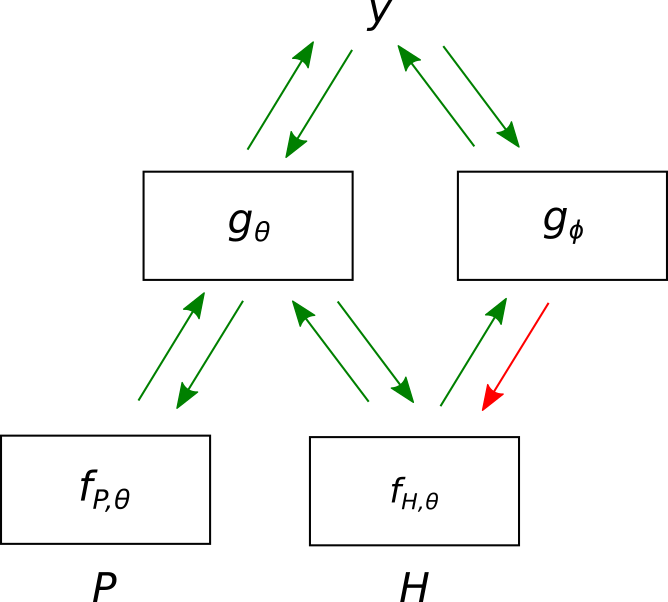}
\caption{\HOClong} 
\label{fig:arch-double}
\end{subfigure} \hfill
\begin{subfigure}[b]{0.32\linewidth}
\centering
\includegraphics[height=3.5cm]{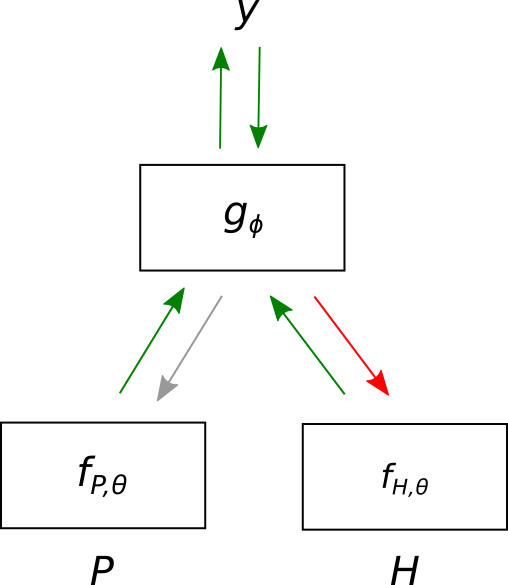}
\caption{\NS} 
\label{fig:arch-single}
\end{subfigure}
\caption{Illustration of (a) the baseline NLI architecture, and
our two proposed methods to remove hypothesis only-biases from
an NLI model:  (b)
uses a hypothesis-only classifier, and (c)
samples
a random premise.
Arrows correspond to the direction of propagation. Green or red arrows respectively mean that the gradient sign is kept as is or reversed. Gray arrow indicates that the gradient is
not back-propagated - this only occurs in (c) when we randomly sample a premise, otherwise the gradient is back-propagated. $f$ and $g$ represent encoders and classifiers.}
\label{fig:archs}
\end{figure*}

Recent studies have tried to create new NLI datasets that do not contain such artifacts, but
many approaches
to dealing with this issue
remain unsatisfactory: constructing new datasets~\citep{sharma-EtAl:2018:Short} is costly and may still result in other artifacts; filtering ``easy'' examples and defining a harder subset is useful for evaluation purposes~\citep{gururangan-EtAl:2018:N18-2}, but difficult to do on a large scale that enables 
training; and 
compiling adversarial examples~\citep{glockner-shwartz-goldberg:2018:Short} is informative but again limited by scale or diversity.
Instead, our goal
is to develop methods that 
overcome these biases as datasets may
still
contain undesired artifacts
despite annotation efforts. 

Typical NLI models learn to predict an entailment label discriminatively given a premise-hypothesis pair (\newfigref{fig:arch-baseline}),
enabling them to learn hypothesis-only biases. Instead,
we 
predict the premise  given the hypothesis and the entailment label, which by design cannot be solved using data artifacts. While this objective is intractable,  it motivates two approximate training methods for standard NLI classifiers that are more resistant to biases.
Our first method uses a hypothesis-only classifier 
(\newfigref{fig:arch-double}) and the second uses negative sampling by swapping premises between premise-hypothesis pairs (\newfigref{fig:arch-single}).

We evaluate  the ability of our methods to generalize better in synthetic and naturalistic settings.
First, using a controlled, synthetic dataset,  we demonstrate that, unlike the baseline, our methods enable a model to ignore the artifacts and learn to correctly identify the desired relationship between the two texts.
Second, we train models on an NLI dataset that is known to be biased and evaluate on other datasets that may have different or no biases. We observe improved results compared to a fully discriminative baseline in $9$ out of $12$ target datasets, indicating that our methods
generate models 
that 
are more robust to annotation artifacts.

An extensive analysis reveals that our methods
are most effective when the target datasets have 
different biases from the source dataset or no noticeable biases.
We also observe that the more we encourage the model to ignore biases, the better it transfers, but this comes at the expense of 
performance on the source dataset.
Finally,
we show that
our methods can better exploit small amounts of training data in a target dataset, especially when it has different biases from the source data.

In this paper, we focus on the transferability of our methods from biased datasets to ones having different or no biases. Elsewhere~\cite{belinkov:2019:starsem}, we have analyzed the effect of these methods on the learned language representations, 
suggesting that they may indeed be less biased. However, we caution that complete removal of biases remains difficult and is dependent on the techniques used. The choice of whether to remove bias also depends on the goal; in an in-domain scenario certain biases may be helpful and should not necessarily be removed.

\medskip

In summary, in this paper we make the following contributions:
\begin{itemize}[itemsep=1pt,topsep=1pt,parsep=1pt,partopsep=1pt]
\item Two novel 
methods to train NLI models that are more robust to dataset-specific artifacts.
\item An empirical evaluation of the methods on a synthetic dataset and $12$ naturalistic 
datasets. 
\item An extensive analysis of the effects of our methods on handling bias.
\end{itemize}

\section{Motivation}
\label{sec:motivation}

A training instance for NLI consists of a hypothesis sentence $H$, a premise statement $P$,
and an inference label $y$. A probabilistic NLI model aims to learn a parameterized
distribution $p_{\theta}(y \mid P, H)$ to compute the probability of the label given the
two sentences. We consider NLI models with premise and hypothesis encoders, $f_{P,\theta}$ and $f_{H,\theta}$, which learn representations of $P$ and $H$, and a classification layer, $g_\theta$, which learns a distribution over $y$. Typically, this is done by maximizing this discriminative likelihood
directly, which will act as our baseline (Figure~\ref{fig:arch-baseline}).

However, many NLI datasets contain biases that allow models to perform non-trivially well when accessing just the
hypotheses~\cite{1804.08117,gururangan-EtAl:2018:N18-2,poliak-EtAl:2018:S18-2}. This allows models to leverage hypothesis-only biases that may be present in a dataset.
A model may perform well on a specific dataset, without identifying whether $\premise$ entails $\hypoth$. \newcite{gururangan-EtAl:2018:N18-2} argue that ``the bulk'' of many models'
``success [is] attribute[d] to the easy examples''.
Consequently, this
may limit how well a model trained on one dataset would perform on other datasets that may have different artifacts. 

Consider an example where $\premise$ and $\hypoth$ are strings from $\{a, b, c\}$,
and an environment where $\premise$ entails $\hypoth$ if and only if the first letters are the same,
as in synthetic dataset A. 
In such a setting, a model should be able to learn the correct condition for $\premise$ to entail $\hypoth$.
\footnote{
This is equivalent to XOR and is learnable by a MLP.
}
\begin{center}
\textbf{Synthetic dataset A}  \\
$(a, a)$ $\rightarrow$ \True \hspace{2em}
$(a, b)$ $\rightarrow$ \False \\
$(b, b)$ $\rightarrow$ \True \hspace{2em}
$(b, a)$ $\rightarrow$ \False \\
\end{center}

Imagine now that an artifact $c$ is appended to every entailed $\hypoth$ (synthetic dataset B).
A model of $y$
with access only to the hypothesis side 
can fit the data perfectly by detecting the presence or absence of $c$ in $\hypoth$,
ignoring the more general pattern.
Therefore, we hypothesize that a model that learns $p_{\theta}(y \mid \premise, \hypoth)$ by training on such data would be misled by the bias $c$ and would fail to learn the relationship between the premise and the hypothesis. Consequently, the model would not perform well on the unbiased synthetic dataset A.  
\begin{center}
\textbf{Synthetic dataset B (with artifact)} \\
$(a, ac)$ $\rightarrow$ \True \hspace{2em}
$(a, b)$ $\rightarrow$ \False \\
$(b, bc)$ $\rightarrow$ \True \hspace{2em}
$(b, a)$ $\rightarrow$ \False \\
\end{center}

Instead of maximizing the discriminative likelihood $p_{\theta}(y \mid P, H)$ directly,
we consider maximizing the likelihood of generating the premise $P$ conditioned
on the hypothesis $H$ and the label $y$: $p(P \mid H, y)$. This objective cannot be fooled
by hypothesis-only features, and it requires taking the premise into account. For example,
a model that only looks for $c$ in the above example cannot do better than chance on this objective. However, as $P$ comes from the space of all sentences, this objective is much more
difficult to estimate.

\section{Training Methods}

Our goal is to maximize $\log p(P \mid H, y)$ on the training data. While we could in theory
directly 
parameterize this distribution, for efficiency and simplicity
we
instead write it in terms of the standard $p_{\theta}(y \mid P, H)$ and introduce a new term
to approximate the normalization:
\[   \log p(P \mid y, H) =  \log\dfrac{p_{\theta}(y
\mid P, H) p(P \mid H)}{p(y \mid H)}. \]
\noindent
Throughout we will assume $p(P \mid H)$ is a fixed constant (
justified by the dataset assumption that, lacking $y$, $P$ and $H$ are independent and drawn at random). Therefore, to approximately maximize
this objective we need to estimate $p(y \mid H)$.
We propose two methods for doing so.

\subsection{\HOClong: Hypothesis-only Classifier} 

Our first approach 
is to estimate the term $p(y \mid H)$ directly. In theory,
if labels in an NLI dataset depend on both premises and hypothesis (which \newcite{poliak-EtAl:2018:S18-2} call ``\textit{interesting} NLI''), this should be a uniform distribution.
However, as 
discussed above, it is often possible to 
correctly predict $y$ based
only on the hypothesis. Intuitively, this model can be interpreted as training a classifier to identify the (latent) 
artifacts in the data.

We define this distribution using a shared representation between our new estimator $p_{\phi,\theta}(y \mid H)$ and $p_{\theta}(y \mid P, H)$. In particular, the two share an embedding of $H$ from the hypothesis encoder $f_{H,\theta}$.
The additional parameters $\phi$ are in the final layer $g_{\phi}$, which we call the \textit{hypothesis-only classifier}. The parameters of this layer $\phi$ are updated to fit $p(y \mid H)$ whereas the rest of the parameters in $\theta$ are updated based on the gradients of  $\log p(P \mid y, H)$.

Training is illustrated in \newfigref{fig:arch-double}.
This interplay is controlled by two hyper-parameters. First,
the negative term is scaled by a hyper-parameter $\alpha$.
Second, the updates of $\enc_\phi$ are weighted by $\beta$.
We therefore minimize the following multitask loss functions (shown for a single example):
\begin{align*}
\max_{\theta } L_1(\theta) &= \log {p_{\theta}(y \mid P, H) } - \alpha \log {p_{\phi,\theta}(y \mid H)} \\
\max_{\phi} L_2(\phi) &=  \beta \log {p_{\phi, \theta}(y \mid H) }
\end{align*}
We implement these together with a gradient reversal layer~\citep{ganin2015unsupervised}.
As illustrated in \newfigref{fig:arch-double}, during back-propagation, we first pass gradients through the hypothesis-only classifier $g_{\phi}$ and then reverse the gradients going to the hypothesis encoder $\enc_{H,\theta}$ (potentially scaling them by $\beta$).
\footnote{This approach may also be seen as adversarial training with respect to the hypothesis, akin to domain-adversarial neural networks~\cite{ganin2016domain}. However, our methods encourage robustness to latent hypothesis biases, without requiring a domain label.}

\subsection{\NSlong: Negative Sampling}

As an alternative to the hypothesis-only classifier, our second method 
attempts to
remove annotation artifacts from the representations by sampling alternative 
premises. 
Consider instead writing the normalization term above as, 
\begin{align*}
-\log p(y \mid H) &= -\log \sum_{P'} p(P' \mid H) p(y \mid P', H) \\
&= -\log {\mathbb E}_{P'} p(y \mid P', H) \\
&\geq - {\mathbb E}_{P'} \log p(y \mid P', H),
\end{align*}
\noindent
where the expectation is uniform and the last step is from Jensen's inequality.\footnote{There are more developed and principled approaches in language modeling for approximating this partition function without having to make this assumption. These include importance sampling \cite{bengio2003quick}, noise-contrastive estimation \cite{gutmann2010noise},
and sublinear partition estimation~\cite{rastogi2015sublinear}.
These are more difficult to apply in the setting of sampling full sentences from an unknown set. We hope to explore methods for applying them in future work.}
As in Method 1, we define a separate $p_{\phi,\theta}(y \mid P', H)$ which shares the embedding layers
from $\theta$, $f_{P,\theta}$ and $f_{H,\theta}$. However, as we are attempting to unlearn hypothesis bias, we block the gradients and do not
let it update the premise encoder $f_{P,\theta}$.\footnote{A reviewer asked about gradient blocking. Our motivation was that, for a random premise, we do not have reliable information to update its encoder. However, future work can explore different configurations of gradient blocking.}
The full setting is shown in \newfigref{fig:arch-single}.

To approximate the expectation, we use uniform samples $P'$ (from other training examples) 
to replace the premise in a ($\premise$, $\hypoth$)-pair, while keeping the label $y$.
We also maximize $p_{\theta, \phi}(y \mid \premise', \hypoth)$ to learn the artifacts in the hypotheses.
We use $\alpha \in [0, 1]$ to control 
the fraction of randomly sampled 
$\premise$'s (so the total number of examples remains the same).
As before, we implement this using gradient reversal scaled by $\beta$.
\begin{align*}
\max_{\theta}  L_1(\theta) &= (1- \alpha)  \log {p_{\theta}(y \mid P, H) } \\
& \hspace{2.6em}  - \alpha \log {p_{\theta, \phi}(y \mid P', H)} \\
\max_{\phi} L_2(\phi) &=  \beta \log {p_{\theta, \phi}(y \mid P', H) }
\end{align*}

Finally,
we share the classifier weights between
$p_{\theta}(y \mid P, H)$ and $p_{\phi,\theta}(y \mid P', H)$.
In a sense this is counter-intuitive,
since $p_{\theta}$ is being trained to unlearn bias, while $p_{\phi,\theta}$ is being trained to learn it.
However, 
if the models are trained separately, they may learn to
co-adapt 
with each other~\cite{elazar2018adversarial}.
If $p_{\phi,\theta}$ is not trained well, we might be fooled to think 
that the representation does not contain any biases, while in fact they are still hidden in the representation. For some evidence that this indeed happens when the models are trained separately, see \citet{belinkov:2019:starsem}.\footnote{ 
A similar situation arises in neural cryptography~\citep{abadi2016}, where an encryptor Alice and a decryptor Bob 
communicate while an adversary Eve tries to eavesdrop on their communication. Alice and Bob are analogous  to the hypothesis embedding and 
$p_{\theta}$, while Eve is analogous to 
$p_{\phi,\theta}$.
In their asymmetric encryption experiments, \citeauthor{abadi2016}
observed seemingly secret communication, which on closer look the adversary was able to eavesdrop on.}

\section{Experimental Setup}

To evaluate how well our methods can overcome hypothesis-only biases, we test our methods on a synthetic dataset
as well as on 
a wide range of existing NLI datasets.
The scenario we aim to address is when training on a source dataset with biases and evaluating on a target dataset with different or no biases. 
We first describe the data and experimental setup before discussing
the results.

\paragraph{Synthetic Data}
We create a synthetic dataset based on 
the motivating example in
Section~\ref{sec:motivation}, where 
$\premise$ entails $\hypoth$ if and only if their first letters are the same. 
The training and test sets have 1K examples each,
uniformly distributed among the possible entailment relations.
In the test set (dataset A), each premise or hypothesis is a single symbol: $\premise, \hypoth \in \{a,b\}$, where $\premise$ entails $\hypoth$ iff $\premise = \hypoth$.
In the training set (dataset B), a letter $c$ is appended to the hypothesis side in the \True examples, but not in the \False examples.
In order to transfer well to the test set, 
a model that is trained on this training set 
needs to learn the underlying relationship---that $\premise$ entails $\hypoth$ if and only if their first letter is identical---rather than relying on the presence of $c$ in the hypothesis side.

\paragraph{Common NLI datasets}
Moving to existing NLI datasets, we train models
on
the Stanford Natural Language Inference dataset~\citep[SNLI;][]{snli:emnlp2015},
since it
is known to contain significant annotation artifacts.
We evaluate the robustness of our methods 
on other, target datasets.

\def\cca#1{\cellcolor{gray!#1}\ifnum #1>100\color{white}\fi{#1}}

\begin{table*}[t]
\centering
\begin{subtable}[b]{0.45\linewidth}
\footnotesize
\centering
\begin{tabular}{l*{7}{c}}
\toprule
& \multicolumn{6}{c}{$\alpha$} \\
\cmidrule(lr){2-7}
$\beta$  & 0.1 & 0.25 & 0.5 & 1 & 2.5 & 5 \\
\cmidrule(lr){1-1}
\cmidrule(lr){2-7}
0.1 & \cca{50} & \cca{50} & \cca{50} & \cca{50} & \cca{50} & \cca{50}\\
0.5 & \cca{50} & \cca{50} & \cca{50} & \cca{50} & \cca{50} & \cca{50}\\
1 & \cca{50} & \cca{50} & \cca{50} & \cca{50} & \cca{50} & \cca{50}\\
1.5 & \cca{50} & \cca{50} & \cca{50} & \cca{50} & \cca{50} & \cca{100}\\
2 & \cca{50} & \cca{50} & \cca{50} & \cca{50} & \cca{100} & \cca{100}\\
2.5 & \cca{50} & \cca{50} & \cca{100} & \cca{75} & \cca{100} & \cca{100}\\
3 & \cca{50} & \cca{100} & \cca{100} & \cca{100} & \cca{100} & \cca{100}\\
3.5 & \cca{100} & \cca{100} & \cca{100} & \cca{100} & \cca{100} & \cca{100}\\
4 & \cca{100} & \cca{100} & \cca{100} & \cca{100} & \cca{100} & \cca{100}\\
5 & \cca{100} & \cca{100} & \cca{100} & \cca{100} & \cca{100} & \cca{100}\\
10 & \cca{100} & \cca{100} & \cca{100} & \cca{100} & \cca{100} & \cca{100}\\
20 & \cca{100} & \cca{100} & \cca{100} & \cca{100} & \cca{100} & \cca{100}\\
\bottomrule
\end{tabular}
\caption{\HOC}
\label{tab:results-toy-a}
\end{subtable}
\begin{subtable}[b]{0.45\linewidth}
\footnotesize
\centering
\begin{tabular}{l*{6}{c}}
\toprule
& \multicolumn{5}{c}{$\alpha$} \\
\cmidrule(lr){2-6}
$\beta$ & 0.1 & 0.25 & 0.5 & 0.75 & 1 \\
\cmidrule(lr){1-1}
\cmidrule(lr){2-6}
0.1 & \cca{50} & \cca{50} & \cca{50} & \cca{50} & \cca{50} \\
0.5 & \cca{50} & \cca{50} & \cca{50} & \cca{50} & \cca{50} \\
1 & \cca{50} & \cca{50} & \cca{50} & \cca{50} & \cca{50} \\
1.5 & \cca{50} & \cca{50} & \cca{50} & \cca{50} & \cca{50} \\
2 & \cca{50} & \cca{50} & \cca{50} & \cca{50} & \cca{50} \\
2.5 & \cca{50} & \cca{50} & \cca{50} & \cca{50} & \cca{50} \\
3 & \cca{50} & \cca{50} & \cca{100} & \cca{50} & \cca{50} \\
3.5 & \cca{50} & \cca{50} & \cca{100} & \cca{50} & \cca{50} \\
4 & \cca{50} & \cca{100} & \cca{100} & \cca{50} & \cca{50} \\
5 & \cca{50} & \cca{50} & \cca{100} & \cca{100} & \cca{50}$^*$ \\
10 & \cca{75} & \cca{100} & \cca{100} & \cca{100} & \cca{50}$^*$ \\
20 & \cca{100} & \cca{100} & \cca{100} & \cca{50}$^*$ & \cca{50}$^*$ \\
\bottomrule
\end{tabular}
\caption{\NS}
\label{tab:results-toy-b}
\end{subtable}
\caption{Accuracies on the synthetic dataset,
when training on the biased training set 
and evaluating on the unbiased test set. 
Darker boxes represent higher accuracies.
$^*$ indicates failure to learn the biased training set; all other configurations learned the training set perfectly.}
\label{tab:results-toy}
\end{table*}

As target datasets, we use the $10$ datasets
investigated by~\citet{poliak-EtAl:2018:S18-2}
in their hypothesis-only study,
plus two test sets: GLUE's diagnostic test set, which was carefully constructed to not contain hypothesis-biases~\citep{wang2018glue}, and  SNLI-hard, a subset of the SNLI test set that is thought to have 
fewer biases~\citep{gururangan-EtAl:2018:N18-2}.
The target datasets include
\textit{human-judged} datasets that used automatic methods to pair premises and hypotheses, and then relied on humans to label the pairs: 
SCITAIL~\citep{SCITAIL}, ADD-ONE-RTE~\citep{P16-1204},
Johns Hopkins Ordinal Commonsense Inference~\citep[JOCI;][]{TACL1082}, Multiple Premise Entailment~\citep[MPE;][]{lai-bisk-hockenmaier:2017:I17-1},\adam{remove MPE and add it to footnote}\yb{I don't think we should remove MPE; I just think we should footnote about the split strategy vs combination}\adam{but what we did makes 0 sense though - when we split $(\{P_{1}... P_{4}\}-H-L)$ into $P_{1}-H-L, P_{2}-H-L$, the label L rarely holds across all 4 new pairs}\yb{I understand, but is it really 0 sense or has some sense? it just feels odd to exclude it all of a sudden. Don't other people do the same by chance? If you feel strongly about it, I'll accept your judgement and let's remove it}\adam{No one would do what we did? Hopefully I'll run the MPE test experiments soon so we can change the numbers.} and Sentences Involving Compositional Knowledge~\citep[SICK;][]{MARELLI14.363.L14-1314}.
The target datasets also include
datasets recast by~\citet{white-EtAl:2017:I17-1} to evaluate different 
semantic phenomena: 
FrameNet+~\citep[FN+;][]{pavlick-EtAl:2015:ACL-IJCNLP2}, Definite Pronoun Resolution~\citep[DPR;][]{rahman-ng:2012:EMNLP-CoNLL}, and Semantic Proto-Roles~\citep[SPR;][]{TACL674}.\footnote{Detailed descriptions of these datasets can be found in \newcite{poliak-EtAl:2018:S18-2}.}
As many of these 
datasets
have different label spaces than SNLI,  we define a mapping (\appref{app:label-mapping}) from our models' predictions to each target dataset's labels.
Finally,
we also test on 
the Multi-genre NLI dataset~\citep[MNLI;][]{N18-1101}, 
a successor to SNLI.\footnote{We leave additional NLI
datasets, such as the Diverse NLI Collection~\cite{poliak2018emnlp-DNC}, for future work.}

\paragraph{Baseline \& Implementation Details}
We use 
\texttt{InferSent}~\citep{D17-1070} as our baseline 
model because it has been shown to work well on popular NLI datasets and is representative of many NLI models.
We use 
separate \mbox{BiLSTM} encoders
to learn
vector representations of $\premise$ and $\hypoth$.\footnote{Many 
NLI models encode $\premise$ and $\hypoth$ separately
~\citep{rocktaschel2015reasoning,mou-EtAl:2016:P16-2,liu2016learning,
cheng-dong-lapata:2016:EMNLP2016,
P17-1152},
although
some share information between the encoders via attention~\citep{D16-1244,attention-fused-deep-matching-network-natural-language-inference}.
}
The vector representations are combined
following~\newcite{mou-EtAl:2016:P16-2},\footnote{Specifically, representations are concatenated, subtracted, and multiplied element-wise.} and passed to an MLP
classifier with one hidden layer.  
Our proposed methods for mitigating biases use the same technique for representing and combining sentences. 
Additional implementation details are provided in \appref{app:implementation}.

For both methods,
we sweep hyper-parameters $\alpha$, $\beta$ over 
$\{0.05, 0.1, 0.2, 0.4, 0.8, 1.0\}$.  
For each target dataset, we choose the best-performing model on its development set and report 
results on the test set.\footnote{For MNLI, since the test sets are not available, we tune on the matched dev set and evaluate on the mismatched dev set, or vice versa. For GLUE, we tune on MNLI matched.}

\newlength\tbspace
\setlength\tbspace{13cm}
\newcolumntype{L}{l<{\hspace{\tbspace}}}

\begin{table*}[t]
\footnotesize
\centering
\begin{tabular}{l  r  H 
H r @{\hspace{2pt}}r@{|}l
H r
@{\hspace{2pt}}r@{|}l   
H  r @{\hspace{2pt}}r@{|}l  H r @{\hspace{2pt}}r@{|}l  }
\toprule
& \multicolumn{10}{c}{Test On Target Dataset} & \multicolumn{8}{c}{Test On SNLI} \\
\cmidrule(lr){2-11}\cmidrule(lr){12-19}
Target Test Dataset & Baseline & Hypothesis & \HOC & \multicolumn{3}{c}{$\Delta$ \HOC}  & \NS & \multicolumn{3}{c}{$\Delta$ \NS}  & \HOC  & \multicolumn{3}{c}{$\Delta$ \HOC} & \NS  & \multicolumn{3}{c}{$\Delta$ \NS} \\
\midrule
SCITAIL  & 58.14 & 37 & 57.67 & -0.47 &  \rule{0.721966206pt}{5pt}  &    & 51.08 & -7.06 &  \rule{2.269366763pt}{5pt}  &    & 84.04 & -0.18 & \rule{0.786026201pt}{5pt} &  & 75.16 & -9.06 & \rule{1.825508765pt}{5pt} & \\
ADD-ONE-RTE  & 66.15 & 40 & 66.15 & 0.00 &    &  \rule{0pt}{5pt}  & 83.46 & 17.31 &    &  \rule{5.56412729pt}{5pt}  & 81.93 & -2.29 & \rule{10pt}{5pt} &  & 34.59 & -49.63 & \rule{10pt}{5pt} & \\
JOCI  & 41.50 & 36 & 41.74 & 0.24 &    &  \rule{0.368663594pt}{5pt}  & 39.63 & -1.87 &  \rule{0.601092896pt}{5pt}  &    & 83.78 & -0.44 & \rule{1.92139738pt}{5pt} &  & 78.3 & -5.92 & \rule{1.192826919pt}{5pt} & \\
MPE  & 57.65 & 34 & 58.1 & 0.45 &    &  \rule{0.69124424pt}{5pt}  & 52.35 & -5.30 &  \rule{1.703632273pt}{5pt}  &    & 83.65 & -0.57 & \rule{2.489082969pt}{5pt} &  & 83.68 & -0.54 & \rule{0.108805158pt}{5pt} & \\
DPR  & 49.86 & 33 & 50.96 & 1.10 &    &  \rule{1.689708141pt}{5pt}  & 49.41 & -0.45 &  \rule{0.144648023pt}{5pt}  &    & 83.49 & -0.73 & \rule{3.187772926pt}{5pt} &  & 76.41 & -7.81 & \rule{1.573644973pt}{5pt} & \\
MNLI matched  & 45.86 & 41 & 47.24 & 1.38 &    &  \rule{2.119815668pt}{5pt}  & 43.76 & -2.10 &  \rule{0.675024108pt}{5pt}  &    & 82.97 & -1.25 & \rule{5.458515284pt}{5pt} &  & 75.29 & -8.93 & \rule{1.79931493pt}{5pt} & \\
FN+  & 50.87 & 28 & 52.48 & 1.61 &    &  \rule{2.47311828pt}{5pt}  & 57.03 & 6.16 &    &  \rule{1.980070717pt}{5pt}  & 82.28 & -1.94 & \rule{8.471615721pt}{5pt} &  & 83.78 & -0.44 & \rule{0.088656055pt}{5pt} & \\
MNLI mismatched  & 47.57 & 42 & 49.24 & 1.67 &    &  \rule{2.565284178pt}{5pt}  & 43.66 & -3.91 &  \rule{1.256830601pt}{5pt}  &    & 82.97 & -1.25 & \rule{5.458515284pt}{5pt} &  & 75.29 & -8.93 & \rule{1.79931493pt}{5pt} & \\
SICK  & 25.64 & 37 & 27.44 & 1.80 &    &  \rule{2.764976959pt}{5pt}  & 56.75 & 31.11 &    &  \rule{10pt}{5pt}  & 83.65 & -0.57 & \rule{2.489082969pt}{5pt} &  & 75.29 & -8.93 & \rule{1.79931493pt}{5pt} & \\
GLUE  & 38.50 &    & 40.49 & 1.99 &    &  \rule{3.056835637pt}{5pt}  & 43.21 & 4.71 &    &  \rule{1.513982642pt}{5pt}  & 82.97 & -1.25 & \rule{5.458515284pt}{5pt} &  & 75.29 & -8.93 & \rule{1.79931493pt}{5pt} & \\
SPR  & 52.48 & 35 & 58.99 & 6.51 &    &  \rule{10pt}{5pt}  & 65.43 & 12.94 &    &  \rule{4.159434266pt}{5pt}  & 82.46 & -1.76 & \rule{7.68558952pt}{5pt} &  & 70.21 & -14.01 & \rule{2.822889381pt}{5pt} & \\
\midrule
SNLI-hard  & 68.02 &    & 66.27 & -1.75 &  \rule{2.688172043pt}{5pt}  &    & 55.6 & -12.42 &  \rule{3.992285439pt}{5pt}  &    &    \multicolumn{3}{c}{} &  &  &  \multicolumn{3}{c}{} \\
\bottomrule
\end{tabular}
\caption{
Accuracy results of transferring representations to new datasets. In all cases the models are trained on SNLI.
Left: 
baseline results on target test sets and differences between the proposed methods and the baseline. Right: 
test results on SNLI with the models that performed best on each target dataset's dev set. $\Delta$ are absolute
differences between the method and the baseline on each target test set (left) or
between the method and the baseline performance ($84.22$)
on SNLI test
(right). Black rectangles show relative changes in each column.
}
\label{tab:results-cross}
\end{table*}

\section{Results}

\subsection{Synthetic Experiments}

To examine how well our methods work in a controlled setup, we train 
on the biased dataset (B), but evaluate 
on the unbiased test set (A).
As expected,
without a method to remove hypothesis-only biases, the baseline
fails to generalize to 
the test set.
Examining its predictions,
we found that the baseline model
learned to rely on the
presence/absence of the bias term
$c$, always predicting \True/\False respectively.

Table \ref{tab:results-toy} shows the results of
our two proposed methods.
As we 
increase the hyper-parameters $\alpha$ and $\beta$,
our methods initially behave like
the baseline, learning the training set but failing on the test set.
However, with strong enough hyper-parameters (moving towards the bottom in the tables),
they perform perfectly on both the biased training set and the unbiased test set. 
For \mbox{\HOC}, stronger hyper-parameters work better.
\NS, in particular, breaks down with too many random samples (increasing $\alpha$), as expected.
We also found that 
\HOC did not require as strong $\beta$ as \NS.
From the synthetic experiments, it seems that \HOC
learns to 
ignore the bias $c$ and learn the desired relationship between $\premise$ and $\hypoth$
across many configurations, 
while \NS requires much stronger $\beta$.

\subsection{Results on existing NLI datasets}

Table~\ref{tab:results-cross} (left block)
reports the results of
our proposed methods
compared to the 
baseline in application to the NLI datasets.  
The method using the hypothesis-only classifier to remove hypothesis-only biases from the model
(\HOC) outperforms the baseline in $9$ out of $12$ target datasets ($\Delta>0$), though most
improvements are small.  The training method using negative sampling  (\NS) only outperforms the baseline in $5$ datasets, $4$ of which are cases where the other method also outperformed the baseline. These gains are much larger than those 
of \HOC. 

We also report results of the proposed methods on the SNLI test set (right block).
As our results improve on the target datasets, we note that \HOC's performance on SNLI does not drastically decrease (small $\Delta$), even when the improvement on the target dataset is large (for example, in SPR). For this method, the performance on SNLI drops by just an average of 1.11 (0.65 STDV). 
For \NS,
there is a large 
decrease
on SNLI 
as results 
drop by an average of 11.19 (12.71 STDV). 
For these models, when we see large 
improvement on a target dataset, we often see a large drop on SNLI. For example, on ADD-ONE-RTE, \NS outperforms the baseline by roughly 17\% 
but performs almost 50\% lower 
on SNLI. 
Based on this, as well as the results on the synthetic dataset, \NS seems to be much more unstable and highly dependent on the right hyper-parameters.

\section{Analysis}
Our results demonstrate that our approaches may be robust
to many datasets with different types of bias.
We next analyze our results and explore modifications to the experimental setup that may improve model transferability 
across NLI datasets.

\subsection{Interplay with known biases}

A priori, we expect
our methods to provide the most benefit
when a target dataset has no hypothesis-only biases or
such biases that differ from ones in the training data.
Previous work estimated the amount of bias in NLI datasets by comparing the performance of a hypothesis-only classifier with the majority baseline~\cite{poliak-EtAl:2018:S18-2}. If the classifier outperforms the baseline, the dataset is said to have hypothesis-only biases.
We follow a similar idea for estimating how similar the biases in a target dataset are to those in the source dataset. 
We compare the performance of a hypothesis-only classifier trained on SNLI and evaluated on each target dataset, 
to a 
majority baseline
of the most frequent class in each target dataset's training set (Maj). 
We also compare 
to a hypothesis-only classifier trained and tested on each target dataset.
\footnote{A reviewer noted that this method may miss similar bias ``types'' that are achieved through different lexical items. We note that our use of pre-trained word embeddings might mitigate this concern. }

\begin{figure}[t]
\centering
\includegraphics[width=1.00\linewidth,trim={0 0.15cm 0 0.9cm},clip]{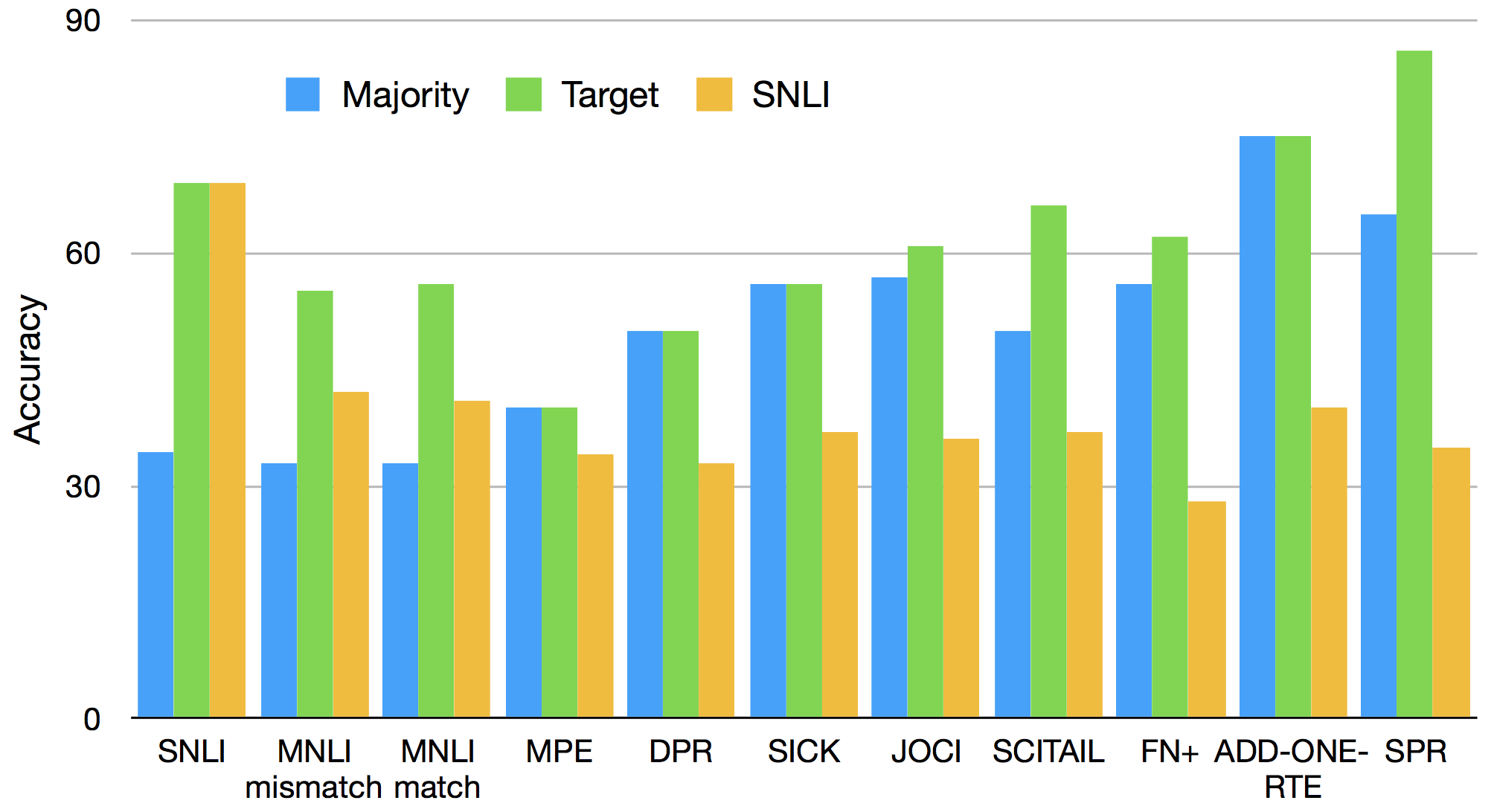}
\caption{Accuracies of majority 
and hypothesis-only baselines on each dataset (x-axis).
The datasets are generally ordered by increasing difference between a hypothesis-only model trained on the target dataset (green) compared to trained on SNLI (yellow).}
\label{fig:hyp-diff-biases}
\end{figure}

\newfigref{fig:hyp-diff-biases} shows the results.
When the hypothesis-only model trained on SNLI is tested on the target datasets, the model performs below Maj (except for MNLI), indicating that these target datasets contain different biases than those in SNLI. 
The largest difference is on SPR: 
a 
hypothesis-only model trained on SNLI performs over 50\% worse than one 
trained on SPR.
Indeed, our methods lead to large improvements on SPR (\tabref{tab:results-cross}), indicating that they are especially helpful when the target dataset contains different biases.
On MNLI, this
hypothesis-only model performs 10\% above Maj,  
and roughly 20\% worse compared to when trained on MNLI, suggesting that MNLI
and SNLI have 
similar biases.
This may explain why our methods 
only
slightly outperform the baseline on MNLI (\tabref{tab:results-cross}). 

\begin{figure*}[t!]
\centering
\begin{subfigure}[t]{0.49\linewidth}
\includegraphics[width=\linewidth]{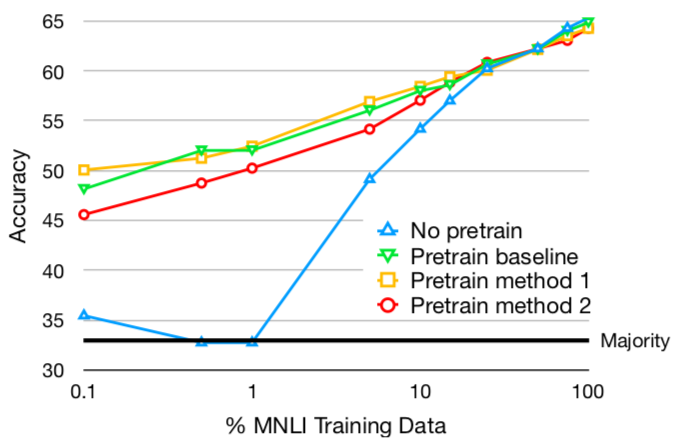}
\end{subfigure} \hfill
\begin{subfigure}[t]{0.49\linewidth}
\includegraphics[width=\linewidth]{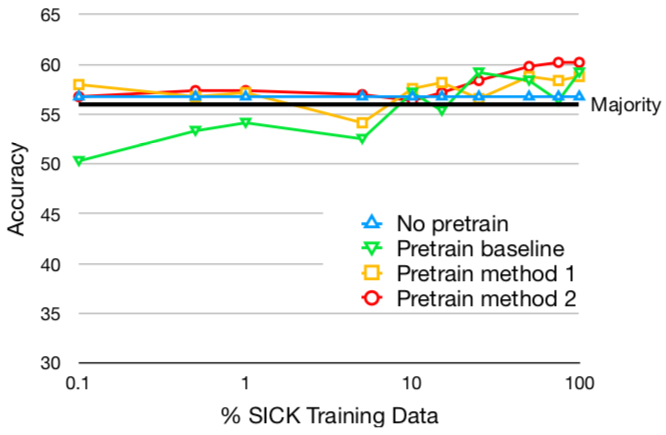}
\end{subfigure}
\caption{Effect of fine-tuning with the baseline and the proposed 
methods on MNLI (left) and SICK (right).}
\label{fig:increasing}
\end{figure*}

The hypothesis-only model
trained on each target dataset 
did not outperform Maj 
on DPR, ADD-ONE-RTE, SICK, and MPE,
suggesting that these datasets do not have noticeable hypothesis-only biases. 
Here, as expected, 
we observe improvements 
when our methods are tested on these datasets, to varying degrees (from 0.45 on MPE to 31.11 on SICK). 
We also see improvements on datasets with biases
(high performance of training on each dataset compared to the corresponding majority baseline), 
most noticeably SPR.
The only exception seems 
to be SCITAIL, where we do not improve despite 
it having different biases than SNLI. However, when we strengthen
$\alpha$ and $\beta$
(below), \HOC outperforms the baseline.

Finally, 
both 
methods obtain improved results
on the GLUE diagnostic
set, 
designed to be bias-free. We do not see improvements on 
SNLI-hard, 
indicating 
it may still have biases --
a possibility acknowledged by 
\citet{gururangan-EtAl:2018:N18-2}.

\subsection{Stronger hyper-parameters}
\label{sec:stronger-hyper-params}

In the synthetic experiment, we found that increasing 
$\alpha$ and $\beta$ improves the
models' ability to generalize to the unbiased dataset. Does the same apply to natural NLI datasets? 
We expect that strengthening the auxiliary losses ($L_{2}$ in our methods) during training will hurt performance on the original data (where biases are useful), but improve on the target data, which may have different or no biases (\newfigref{fig:hyp-diff-biases}). To test this,
we increase the hyper-parameter values during training;
we consider the range $\{ 1.5, 2.0, 2.5, 3.0, 3.5, 4.0, 4.5, 5.0\}$.
\footnote{
The synthetic setup 
required
very strong hyper-parameters, possibly due to the clear-cut nature of the task.  
In 
the
natural NLI setting,
moderately strong 
values 
sufficed.
}
While there are other ways to
strengthen our methods, 
e.g., increasing the number or size of hidden layers~\citep{elazar2018adversarial}, we are 
interested in the effect of $\alpha$ and $\beta$ as they control 
how much bias is subtracted from our baseline model.

Table~\ref{tab:results-cross-stronger} shows the results of \HOC with stronger hyper-parameters
on the existing NLI datasets. 
As expected, performance on SNLI test sets (SNLI and SNLI-hard in~\tabref{tab:results-cross-stronger}) decreases more, but many of the other datasets benefit from stronger hyper-parameters 
(compared with~\tabref{tab:results-cross}). We see the largest improvement on SICK, achieving 
over 10\% 
increase 
compared to the 1.8\% 
gain in~\tabref{tab:results-cross}.
As for \NS, 
we found  large drops in quality even in our basic configurations 
(\appref{app:cv}), so we do not increase the hyper-parameters further.
This should not be too surprising, 
adding too many 
random premises 
will lead to
a model's degradation. 

\begin{table}[t]
\footnotesize
\centering
\begin{tabular}{lccr @{\hspace{2pt}}r@{|}l} 
\toprule
Dataset & \multicolumn{1}{c}{Base} & \multicolumn{1}{c}{\HOC} & \multicolumn{3}{c}{$\Delta$} \\ 
\midrule
JOCI & 41.50 & 39.29 & -2.21 & \rule{2.01826484pt}{5pt} &  \\
SNLI & 84.22 & 82.40 & -1.82 & \rule{1.662100457pt}{5pt} &  \\
DPR & 49.86 & 49.41 & -0.45 & \rule{0.410958904pt}{5pt} &  \\
MNLI matched & 45.86 & 46.12 & 0.26 &  & \rule{0.237442922pt}{5pt} \\
MNLI mismatched & 47.57 & 48.19 & 0.62 &  & \rule{0.566210046pt}{5pt} \\
MPE & 57.65 & 58.60 & 0.95 &  & \rule{0.867579909pt}{5pt} \\
SCITAIL & 58.14 & 60.82 & 2.68 &  & \rule{2.447488584pt}{5pt}\\
ADD-ONE-RTE & 66.15 & 68.99 & 2.84 &  & \rule{2.593607306pt}{5pt}\\
GLUE & 38.50 & 41.58 & 3.08 &  & \rule{2.812785388pt}{5pt}\\
FN+ & 50.87 & 56.31 & 5.44 &  & \rule{4.96803653pt}{5pt}\\
SPR & 52.48 & 58.68 & 6.20 &  & \rule{5.662100457pt}{5pt}\\
SICK & 25.64 & 36.59 & 10.95 &  & \rule{10pt}{5pt}\\
\midrule
SNLI-hard & 68.02 & 63.81 & -4.21 & \rule{3.844748858pt}{5pt} \\
\bottomrule
\end{tabular}
\caption{Results with stronger hyper-parameters for \HOC 
vs.\ 
the
baseline. $\Delta$'s are absolute differences.  
}
\label{tab:results-cross-stronger}
\end{table}

\subsection{Fine-tuning on target datasets}

Our main goal is to determine whether our 
methods
help a model perform 
well across multiple datasets by ignoring dataset-specific 
artifacts. In turn, we did not update the models' parameters on other datasets. 
But, what
if we are given different amounts of training data for a new NLI dataset? 

To determine if our approach is still helpful, we updated four 
models on increasing sizes of training data from two target datasets (
MNLI and SICK).
All three training approaches---
the baseline, 
\HOC, 
and 
\NS---are 
used to pre-train a model on SNLI and fine-tune 
on the target dataset. The fourth model is 
the baseline trained only 
on the target dataset.
Both MNLI and SICK have the same label spaces as SNLI, allowing us to hold that variable constant.
We use SICK because our methods resulted in good gains on it (\tabref{tab:results-cross}).
MNLI's large training set
allows us to consider a wide range of
training set sizes.\footnote{
We hold out $10$K examples from the training set for dev as gold labels for the MNLI test set are not publicly available.
We evaluate on MNLI's matched dev set to assure consistent domains when fine-tuning.}

\newfigref{fig:increasing} shows the results on the dev sets. In 
MNLI, 
pre-training is very helpful when fine-tuning on a small amount of new training data, although there is little to no gain from pre-training with either of our methods compared to the baseline.
This is expected, as we saw relatively small gains with the proposed methods on MNLI, and can be explained by SNLI and MNLI having similar biases.
In 
SICK, 
pre-training with either of our methods is
better in most data regimes,
especially with very small amounts of target training data.\footnote{Note that SICK is a small dataset
(10K 
examples), which explains why the
model without
pre-training does not benefit from more data, barely surpassing the majority baseline.  
} 
25\% of the 

\section{Related Work} \label{sec:related work}
\paragraph{Biases and artifacts in NLU datasets}
Many natural language undersrtanding (NLU) 
datasets contain annotation artifacts. Early work on NLI, also known as recognizing textual entailment (RTE), found biases that allowed  models to perform relatively well by focusing on syntactic
clues alone~\citep{N06-1005,vanderwende2006syntax}.
Recent work also found artifacts in new NLI datasets~\cite{1804.08117,gururangan-EtAl:2018:N18-2,poliak-EtAl:2018:S18-2}.

Other NLU datasets also exhibit biases. 
In ROC Stories~\citep{mostafazadeh-EtAl:2016:N16-1}, a story cloze 
dataset,
\citet{schwartz2017effect} obtained a high performance by only considering the candidate endings,
without even looking at the story context.  
In this case,
stylistic features of the candidate endings alone, such as the length or 
certain words,
were 
strong indicators of the correct ending
~\citep{K17-1004,P17-2097}. A similar phenomenon was observed in reading comprehension,
where systems performed non-trivially well by  using only the final sentence in the passage or  ignoring the passage altogether~\citep{kaushik2018much}.
Finally, multiple studies found non-trivial performance in visual question answering (VQA) by using only the question, without 
access to the image, due to question biases 
\citep{7780911,7780907,kafle2017visual,goyal2017making,agrawal2017don}.

\paragraph{Transferability across NLI datasets}
It has been known that many NLI models do not transfer
across NLI datasets. Chen Zhang's thesis~\cite{Zhang:2010:NLI:2019860} focused on this phenomena as he demonstrated that ``techniques developed for textual entailment`` datasets, e.g., RTE-3, do not transfer well to other domains, specifically \textit{conversational entailment}~\cite{Zhang:2009:WKC:1708376.1708406,zhang-chai-2010-towards}. \newcite{snli:emnlp2015} and \newcite{N18-1101}
demonstrated (specifically in their respective Tables 7 and 4) how 
models trained on SNLI and MNLI may not transfer well across
other NLI datasets like SICK. 
\newcite{DBLP:journals/corr/abs-1810-09774} recently reported similar findings
using
many advanced deep-learning models.

\paragraph{Improving model robustness}
Neural networks are 
sensitive to adversarial examples, primarily in 
machine vision, 
but also in NLP~\citep{jia-liang:2017:EMNLP2017,belinkov2018synthetic,C18-1055,heigold2017robust,P18-1176,P18-1079,belinkov2019analysis}.
A common approach to improving
robustness is to 
include adversarial examples in 
training~\citep{szegedy2013intriguing,goodfellow2014explaining}. However, this 
may not generalize well to new types of 
examples~\citep{yuan2017adversarial,tramèr2018ensemble}.

Domain-adversarial neural networks aim to increase robustness to domain change, by learning to be oblivious to the domain using gradient reversals~\citep{ganin2016domain}. Our methods rely similarly on gradient reversals when encouraging models to ignore dataset-specific artifacts.
One distinction is that domain-adversarial networks 
require knowledge of the domain at training time,
while our methods
learn to ignore latent 
artifacts 
and do not require direct supervision in the form of a domain label.

Others have attempted to remove biases from learned representations,
e.g., gender biases in word embeddings~\citep{bolukbasi2016man} or
sensitive information like sex and
age in text representations~\citep{P18-2005}.
However, 
removing such attributes 
from text representations may be difficult~\citep{elazar2018adversarial}.
In contrast to this line of work, our final goal is not the removal of such attributes per se; instead, we strive for more robust representations that better transfer to other datasets, similar to \citet{P18-2005}. 

Recent work has applied 
adversarial learning to NLI. 
\newcite{minerviniconll18} generate adversarial examples that do not conform to logical rules and 
regularize models based on those examples. Similarly, \newcite{kang-EtAl:2018:Long} incorporate external linguistic
resources and use a GAN-style framework
to adversarially train robust NLI models.
In contrast, 
we do not use external resources and we are interested in mitigating 
hypothesis-only biases.
Finally, a similar approach has recently been used
to mitigate biases in VQA~\citep{ramakrishnan2018overcoming,grand:2019:SIVL}.

\section{Conclusion} 

Biases in annotations are a major source of concern for the quality of NLI datasets and systems.
We presented a solution for combating annotation biases 
by proposing two training methods to predict the probability of a premise given an entailment label and a hypothesis. 
We demonstrated that this
discourages the hypothesis encoder from learning the biases to instead obtain a less biased representation.
When empirically evaluating our approaches,
we found that in a synthetic setting, as well as on a wide-range of existing NLI datasets, our methods perform better than the traditional training method to 
predict a label given a premise-hypothesis pair.
Furthermore,
we performed several analyses into the interplay of our methods with known biases in NLI datasets,
the effects of stronger bias removal, 
and the possibility of
fine-tuning on the target datasets.

Our methodology can be extended to handle biases in other 
tasks 
where one is concerned with finding  relationships between two objects, such as  visual question answering,   story cloze completion, and reading comprehension.
We hope to encourage such investigation in the broader 
community. 

\section*{Acknowledgements}
We would like to thank Aviad Rubinstein and Cynthia Dwork for discussing an earlier version of this work and the anonymous reviewers for their useful comments. Y.B.\ was supported by the Harvard Mind, Brain, and Behavior Initiative.
A.P. and B.V.D were supported by JHU-HLTCOE and
DARPA LORELEI.
A.M.R gratefully acknowledges the support of NSF 1845664. Views and conclusions contained in this publication are those of the
authors and should not be interpreted as representing official policies or endorsements of DARPA or
the U.S.\ Government.

\bibliography{references}
\bibliographystyle{iclr2019_conference}

\clearpage

\appendix

\section{Appendix}

\subsection{Mapping labels}
\label{app:label-mapping}
Each premise-hypothesis pair in SNLI is labeled  as 
\textsc{entailment}, \textsc{neutral}, or \textsc{contradiction}. MNLI, SICK, and MPE use the same label space. Examples in JOCI are labeled on a 5-way ordinal scale. We follow \newcite{poliak-EtAl:2018:S18-2} by converting it ``into
3-way NLI tags where 1 maps to \textsc{contradiction},
2-4 maps to \textsc{neutral}, and 5 maps to \textsc{entailment}.'' Since examples in SCITAIL are labeled as \textsc{entailment} or \textsc{neutral}, when  evaluating on SCITAIL, we convert the model's  \textsc{contradiction} to \textsc{neutral}. ADD-ONE-RTE and the recast datasets also model NLI as a binary prediction task. However, their label sets are \textsc{entailed} and \textsc{not-entailed}. In these cases, when the models predict \textsc{entailment}, we map the label to \textsc{entailed}, and when the models predict \textsc{neutral} or \textsc{contradiction}, we map the label to \textsc{not-entailed}.

\subsection{Implementation details} \label{app:implementation}

For our experiments on the synthetic dataset, the characters are embedded with 10-dimensional vectors.
Input strings are represented as a sum of character embeddings, and the premise-hypothesis pair is represented by
a concatenation of these embeddings. The classifiers are single-layer MLPs of size 20 dimensions.
We train these models with SGD until convergence.
For the traditional NLI datasets, we use pre-computed 300-dimensional GloVe embeddings~\cite{pennington2014glove}
.\footnote{Specifically,
glove.840B.300d.zip.} The sentence representations learned by the \mbox{BiLSTM} encoders
and the MLP classifier's hidden layer
have a dimensionality of 2048 and 512 respectively.
We follow \texttt{InferSent}'s training regime, using SGD with an initial learning rate of 0.1
and optional early stopping. See \citet{D17-1070} for details.

\subsection{Hyper-parameter sweeps} \label{app:cv}

\begin{figure}[t!]
\centering
\begin{subfigure}[t]{\linewidth}
\includegraphics[width=\linewidth]{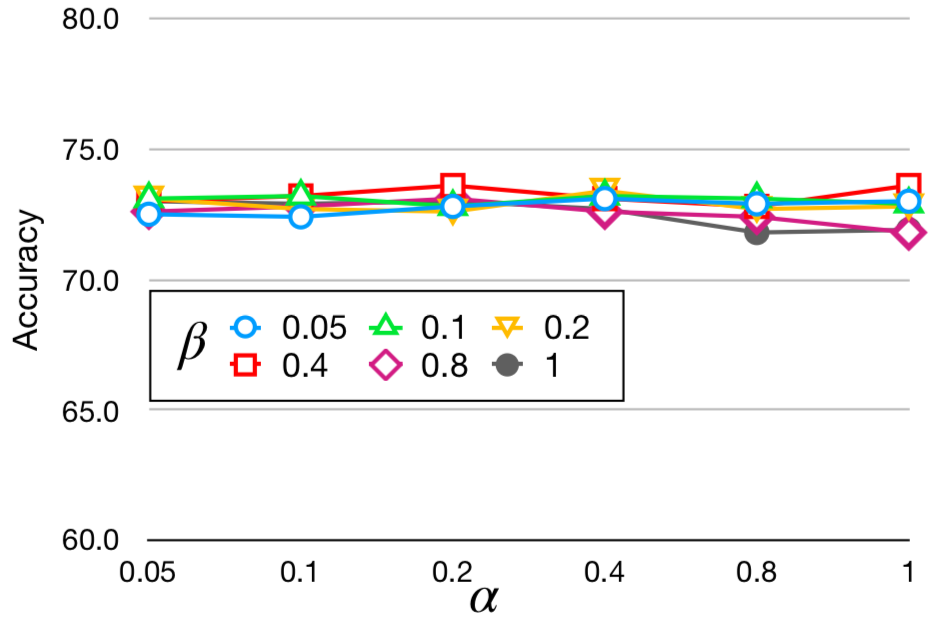}
\caption{\HOC}
\label{fig:app-cv-double}
\vspace{1.5em}
\end{subfigure} \\ 
\begin{subfigure}[t]{\linewidth}
\centering
\includegraphics[width=\linewidth]{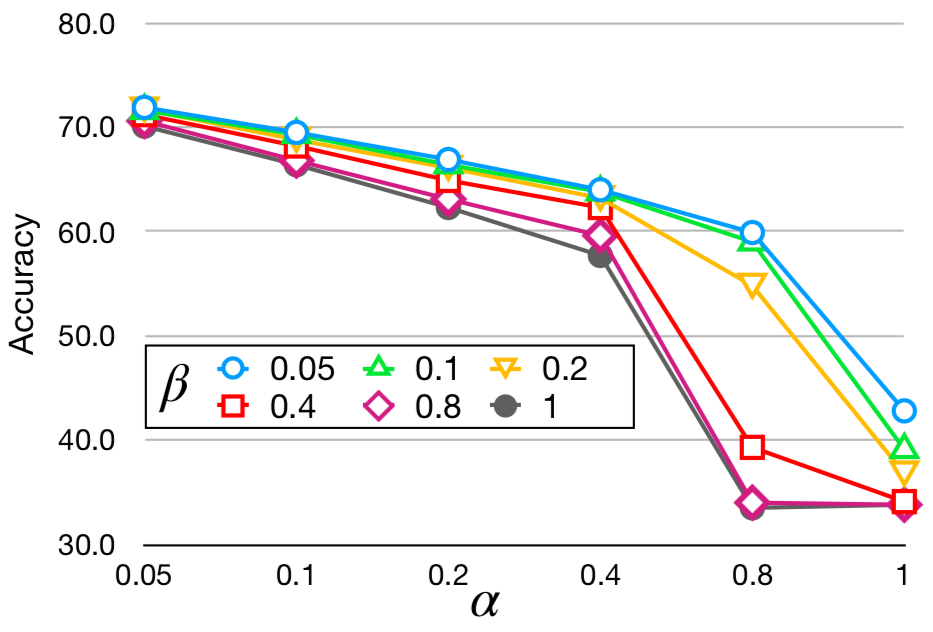}
\caption{\NS}
\label{fig:app-cv-single}
\end{subfigure}
\caption{Cross-validation results.}
\label{fig:app-cv}
\end{figure}

Here we provide 10-fold cross-validation results on a subset of the SNLI training data (50K sentences) with different settings of our hyper-parameters.
\newfigref{fig:app-cv-single} shows the dev set results with different configurations of \NS. 
Notice that performance degrades quickly when we increase the fraction of random premises (large $\alpha$). 
In contrast, the results with \HOC 
(\newfigref{fig:app-cv-double}) are more stable.

\vfill\null
\columnbreak

\end{document}

%% file: math_commands.tex
\usepackage{amsmath,amsfonts,bm}

\def\eqref#1{equation~\ref{#1}}

\def\1{\bm{1}}

\DeclareMathAlphabet{\mathsfit}{\encodingdefault}{\sfdefault}{m}{sl}
\SetMathAlphabet{\mathsfit}{bold}{\encodingdefault}{\sfdefault}{bx}{n}

